\documentclass[conference, 10pt]{IEEEtran}

\usepackage{amsfonts}
\usepackage{xcolor,soul,framed} 

\usepackage[left=1.62cm,right=1.62cm,top=1.9cm, bottom=1in]{geometry}

\colorlet{shadecolor}{yellow}
\usepackage[pdftex]{graphicx}
\graphicspath{{../pdf/}{../jpeg/}}
\DeclareGraphicsExtensions{.pdf,.jpeg,.png}

\usepackage[cmex10]{amsmath}

\usepackage{array}
\usepackage{mdwmath}
\usepackage{amsmath}
\usepackage{amssymb}
\usepackage{mathtools}
\usepackage{mathrsfs}
\usepackage{mdwtab}
\usepackage{eqparbox}
\usepackage{url}
\usepackage{blkarray}
\usepackage{newtxtext,newtxmath}

\usepackage{atbegshi} 
\newlength{\origcolumnsep}

\usepackage[binary-units,per-mode=symbol]{siunitx}
\usepackage[linesnumbered,ruled,vlined,noend]{algorithm2e}

\usepackage{xcolor}

\SetCommentSty{mycommfont}
\SetKwInput{KwInput}{Input}
\SetKwInput{KwOutput}{Output}
\SetKwInput{KwRequire}{Require}
\SetKwInput{KwEnsure}{Ensure}
\SetKw{KwRet}{return}

\usepackage{booktabs}

\usepackage{algpseudocode} 
\usepackage{cuted}
\usepackage{balance} 
\usepackage{etoolbox}

\DeclareMathOperator{\p}{\mathbb{P}}

\begin{document}
 \bstctlcite{IEEEexample:BSTcontrol}
     \title{Adaptive Client Selection via Q-Learning-based Whittle Index in Wireless Federated Learning}


\newcommand{\timothy}[1]{\textcolor{cyan}{[#1]}}
\newcommand{\qiyue}[1]{\textcolor{magenta}{[#1]}}


\author{\IEEEauthorblockN{Qiyue Li\IEEEauthorrefmark{1}, Yingxin Liu\IEEEauthorrefmark{2}, Hang Qi\IEEEauthorrefmark{1}, Jieping Luo\IEEEauthorrefmark{3},  Zhizhang Liu\IEEEauthorrefmark{1}, Jingjin Wu\IEEEauthorrefmark{1}\\}
\IEEEauthorblockA{\IEEEauthorrefmark{1}Guangdong Prov./Zhuhai Key Lab of IRADS, Beijing Normal-Hong Kong Baptist University, Guangdong, P. R. China\\}
\IEEEauthorblockA{\IEEEauthorrefmark{2}College of Engineering, Carnegie Mellon University,
Pittsburgh, PA, USA \\}
\IEEEauthorblockA{\IEEEauthorrefmark{3}Department of Statistics, University of Oxford, 24-29 St Giles', Oxford OX1 3LB, U.K.\\}
Email: t330005026@mail.uic.edu.cn; yingxin2@andrew.cmu.edu; t330202706@mail.uic.edu.cn; 
\\
jieping.luo@reuben.ox.ac.uk; t330031181@mail.uic.edu.cn; jj.wu@ieee.org\\}

\maketitle

\begin{abstract}
We consider the client selection problem in wireless Federated Learning (FL), with the objective of reducing the total required time to achieve a certain level of learning accuracy. Since the server cannot observe the clients' dynamic states that can change their computation and communication efficiency, we formulate client selection as a restless multi‑armed bandit problem. We propose a scalable and efficient approach called the Whittle Index Learning in Federated Q-learning (WILF-Q), which uses Q-learning to adaptively learn and update an approximated Whittle index associated with each client, and then selects the clients with the highest indices. Compared to existing approaches, WILF-Q does not require explicit knowledge of client state transitions or data distributions, making it well-suited for deployment in practical FL settings. Experiment results demonstrate that WILF-Q significantly outperforms existing baseline policies in terms of learning efficiency, providing a robust and efficient approach to client selection in wireless FL.

\end{abstract}

\begin{IEEEkeywords}
Federated learning, client selection, Q-learning, index policy 
\end{IEEEkeywords}

%
\IEEEpeerreviewmaketitle


\section{Introduction}


Federated Learning (FL) is a distributed machine learning paradigm that allows multiple clients to collaboratively train a shared model while keeping their data private. FL enables clients to perform training tasks locally and share only model updates, which are aggregated to update the global model at the server by decentralizing the learning process and keeping potentially sensitive raw data locally. FL has been adopted in various applications with privacy preservation requirements, including medical diagnostics, satellite communications, and real-time analytics in Internet of Things~\cite{yang_federated_2019,qi2025energy}. 

Client selection in FL refers to the process of determining which participating devices (clients) contribute to each training round. Intuitively, it directly influences the overall training efficiency and convergence performance. This importance arises primarily because FL clients are inherently heterogeneous, exhibiting significant variability in data distribution and computational capabilities. Such heterogeneity becomes even more significant in dynamic environments like edge computing, where client-specific factors, including energy availability and network connectivity, may fluctuate considerably over time, thereby introducing additional complexity into client selection decisions~\cite{fu_client_2023}. 

Additionally, unlike centralized systems with static and fully observable conditions, wireless FL operates within dynamically evolving environments where client states related to FL operation can change rapidly~\cite{nishio_client_2019}. For example, in edge computing contexts, the potential clients include battery-powered wearable health devices that may experience battery depletion or poor network connectivity. Selecting clients in poor condition may result in longer training or communication time, leading to lower overall learning efficiency. 

While many existing studies on FL client selection policies (e.g.,~\cite{abdulrahman_fedmccs_2021}) addressed non-independent and identically distributed (non-IID) data among clients, the impact of dynamically evolving client states has not been adequately considered. Typically, these approaches assume static client states, implying that each client's computational and communication capacities remain constant over time. Under such assumptions, resource-aware client selection strategies have been demonstrated to perform satisfactorily, even under non-IID conditions. 

However, the above-mentioned static-state methods do not address variability issues of clients. Notably, given the limited battery capacities of mobile clients in wireless FL, client availability states often exhibit a strong dependency on their recent selection history. As local training tasks are generally computationally intensive for mobile devices, a client recently selected for participation will likely experience decreased availability in subsequent rounds due to resource exhaustion~\cite{Arouj2022}.

This scenario aligns naturally with the framework of restless multi-armed bandit problems (RMABP), which generalize traditional multi-armed bandit approaches by explicitly incorporating evolving states. A distinguishing feature of RMABP compared to standard multi-armed bandit formulations is that the state of each client continues to evolve regardless of whether it is actively selected. Thus, in wireless FL, the condition of a client, whether selected or not, can change dynamically from round to round.

The Whittle index policy provides a computationally feasible heuristic to schedule clients that approximately maximizes the long-term learning reward in generally NP-hard RMABPs~\cite{Whittle1988}. However, one potential obstacle to applying Whittle index in a wireless FL scenario is that the transition dynamics of clients are not fully observable to the server coordinating the training~\cite{albelaihi_green_2022}. 

In this paper, we apply Q-learning to learn the approximate Whittle index for client selection in wireless FL. The main advantage of Q-learning is that it does not require explicit knowledge of state transition probabilities or reward distributions. This is particularly valuable in FL, where precise dynamics are often unknown or difficult to model accurately. In addition, Q-learning can adaptively favor clients that contributed to faster convergence, by accounting for their historical accuracy gains when selected. While Q-learning has been demonstrated to effectively address the Age of Information issue in FL, by identifying the clients with fresh datasets through approximated Whittle index~\cite{wu_towards_2023}, our approach learns an index that reflects each client’s current computing throughput, uplink rate, and contribution to improving accuracy. By targeting these conditions, our method adapts to fluctuations in battery level, CPU load, and channel quality—factors that largely determine efficiency in resource‑constrained wireless FL environments. This shifts the focus from data timeliness as considered in most existing studies adopting Q‑learning–based index policies to client capability (e.g.,~\cite{wu_towards_2023}), a dimension that have not been sufficiently explored.

Our main contributions are summarized as follows.

\begin{itemize}
\item We present a practically viable solution for client selection in wireless FL systems where dynamic states and potential contributions in accuracy gains of clients are not fully observable by the server. We employ Q-learning to learn the approximated Whittle indices of clients, overcoming the complexities and uncertainties encountered by battery-powered edge devices. 

\item We propose an adaptive algorithm called the Whittle Index Learning in Federated Q-learning (WILF-Q), designed to achieve scalable and robust client selection, thereby enhancing overall training efficiency. Numerical experiments show that WILF-Q can reduce the training time by up to 45\% compared to baseline methods including a recently proposed multi-armed bandit-based approach~\cite{Chen2022heterogeneous}.

\item Our approach significantly extends prior research by integrating Whittle Index theory and Q-learning into a cohesive framework for optimal client selection in wireless FL scenarios with dynamic client states. Although previous studies have applied Whittle Index-based policies to select FL clients based on dataset freshness~\cite{wu_towards_2023}, or Q-learning for optimal resource allocation in FL~\cite{maciel_federated_2024}, our work  simultaneously captures dynamic state variations of clients and their evolving impact on training effectiveness. 

\end{itemize}




\section{System Model}

We consider a wireless network, where $\mathcal{C}_r$ denotes the set of available federated learning (FL) clients for potential selection in training round $r$, and $\mathbf{w}^r$ denotes the global parameters at the end of that round. Due to certain communication constraints, only a subset of clients, $\mathcal{A}_r \subset \mathcal{C}_r$, is allowed to participate in each round of training.



The total time required per FL round consists of two components: (local) training time and uplink transmission time. The expected local training time for a selected client $j \in \mathcal{A}_r$ at  $r$, denoted as $t^t_{j,r}$, depends on both its computational capacity and current state. Following standard modeling in distributed systems, we represent $t^t_{j,r}$ as a shifted exponential distribution~\cite{Lee2018Speeding}:
\begin{equation}
        \begin{aligned}
           &\p [t^t_{j,r} <t] = \left\{ 
           \begin{aligned}
               &1-e^{-\frac{s_{j,r}}{D_j}(t-c_{j}D_j)},  \quad  t \ge c_{j}D_j, \\
               &0 , \quad  \text{otherwise,}
           \end{aligned}
           \right. 
         \end{aligned}
\end{equation}
where $c_{j}$ is the computational capacity of client $j$, $D_j$ is the number of local training samples, and $s_{j,r}$ is a coefficient that reflects the state of client $j$ in round $r$.

For analytical tractability, we assume each client could be in one of three states per round, namely normal, limited, or busy. Correspondingly, we have the general condition regarding $s_{j,r}$ values as $0<s^{\text{normal}}_{j,r} < s^{\text{limited}}_{j,r} < s^{\text{busy}}_{j,r} <\infty$ for any given pairs of $j$ and $r$. The exact states of clients are not observable by the server but are assumed to follow one in a pair of discrete-time Markov processes, depending on whether the client is selected or not in the current round. Each client belongs to a class $c \in \mathcal{C}$, where all clients in the same class share identical transition dynamics.
Let $S=\{n,l,b\}$ denote the state space with $n$, $l$, and $b$ corresponding to normal, limited, and busy states, respectively. The transition probability from state $x$ to $y$ is denoted $P^{c}_s(x,y)$ when the client is selected, and $P^{c}_n(x,y)$ otherwise. 
This transition model captures realistic wireless behavior: selected clients tend to transition toward resource-constrained states (e.g., from normal to busy), due to power and computation consumption, while unselected clients are more likely to recover to the normal state.
A sample of the transition matrices for a given client class $c$ is demonstrated as follows.
\[
\renewcommand{\arraystretch}{1.4} 
P^{c}_s = 
\begin{blockarray}{cccc}
n & l & b  \\
\begin{block}{(ccc)c}
  \frac{1}{2} & \frac{1}{3} & \frac{1}{6} & n\\
  \frac{1}{6} & \frac{1}{2} & \frac{1}{3} & l\\
  \frac{1}{6} & \frac{1}{6} & \frac{2}{3} & b\\
\end{block}
\end{blockarray}
\quad
\text{and}
\quad
P^{c}_n = 
\begin{blockarray}{cccc}
n & l & b  \\
\begin{block}{(ccc)c}
  \frac{2}{3} & \frac{1}{6} & \frac{1}{6} & n\\
  \frac{1}{3} & \frac{1}{2} & \frac{1}{6} & l\\
  \frac{1}{6} & \frac{1}{3} & \frac{1}{2} & b\\
\end{block}
\end{blockarray}
\]

The uplink communication time $t^c_{j,r}$ for client $j$ in round $r$ is
\begin{equation}
t^c_{j,r} = \frac{U_j}{B_j\log_2(1+\frac{p_j g_j^2}{\sigma_0})},
\label{eq:com_time}
\end{equation}
where $U_j$ is the model parameter size, $B_j$ is the uplink bandwidth, $p_j$ is the transmission power, $\sigma_0$ is the thermal noise, and $g_j^2$ is the  instantaneous channel gain which follows an exponential distribution with mean$\mu_j$~\cite{Zhao2022System}. 

To ensure bounded training duration, the FL system imposes  a maximum per-round latency $t_\text{max}$; clients exceeding this threshold are excluded from model aggregation. The round latency is therefore given by
\begin{equation}
T^\phi_{r} = \min(t_\text{max},\max_{j \in \mathcal{A}_r} (t^t_{j,r}+t^c_{j,r})).
\end{equation}
which reflects the latency bottleneck induced by the slowest selected client.

During each round \(r\), every selected client \(j \in \mathcal{A}_r\) downloads previous global model $\mathbf{w}^{r-1}$, performs local updates, and obtains a personalized local model $\mathbf{w}^r_j$. The server then aggregates these local models using data-proportional weights:
\begin{equation}
\mathbf{w}^{r} = \sum_{j \in \mathcal{A}_r} k_j \mathbf{w}_j^{r}.
\end{equation}
where $k_j$ is the weight proportional to client $j$'s local dataset size, satisfying $\sum_j k_j = 1$.

The local loss function $F_j(\mathbf{w})$ is defined as the empirical risk over client $j$'s dataset. The global loss function $F(\mathbf{w})$ is the weighted sum over all clients:
\begin{equation}
F(\mathbf{w}) = \sum_{j \in \mathcal{C}} k_j F_j(\mathbf{w}),
\end{equation}
which quantifies the population-wide model performance.

Since in each round clients locally update $\mathbf{w}$, the global loss is evaluated using their personalized models:
\begin{equation}
F(\mathbf{w}^r) = \sum_{j \in \mathcal{A}_r} k_j F_j\left(\mathbf{w}_j^r\right),
\label{eq:global_loss}
\end{equation}
which approximates the true global loss conditioned on the selected clients and their local updates.

The objective is to minimize the expected total time required to achieve a high level of model accuracy (or equivalently, low loss). Specifically, we aim to design a client selection policy \(\phi \in \Phi\) that minimizes the cumulative  expected latency needed to reduce the expected global loss below a certain threshold $\alpha \in (0,1)$. Formally, the optimization problem is stated as:
\begin{equation}
\label{eq:problem}
\min_{\phi \in \Phi} \mathcal{T}^\phi = \sum_{r=1}^{R^\phi} \mathbb{E}[T^\phi_r], \quad \text{s.t.} \quad 
\mathbb{E}[F(\mathbf{w}^{(R^\phi)}) - F(\mathbf{w}^*)] \leq \alpha,
\end{equation}
where \(R^\phi\) denotes the number of rounds needed under policy \(\phi\) to reach the accuracy threshold $\alpha$, \(\mathbb{E}[T^\phi_r]\) is the expected per-round latency, and $\mathbf{w}^* =\arg\min_{\mathbf{w} }  F(\mathbf{w})$ is the optimal parameter minimizing the global loss function.

This is an instance of the \textit{Restless Multi-Armed Bandit Problem (RMABP)}, as states of clients would possibly change regardless of whether they are selected or not. Such problem has been shown to be PSPACE-hard~\cite{Papadimitriou1994complexity}, as the joint state space grows exponentially with the number of clients.





\section{The WILF-Q Approach}

To overcome the complexity issue in solving~\eqref{eq:problem}, we propose the Whittle
Index Learning in Federated Q-learning (WILF-Q). In each round, the policy selects the top $|\mathcal{A}_r|$ clients with highest estimated indices. Such Whittle index-based policy has been demonstrated effective in RMABP scenarios~\cite{Whittle1988}. However, since the state transition dynamics of clients are unknown to the server, direct  computation of the indices is infeasible. We therefore adopt the Q-learning approach to learn the (unknown) Whittle Indices along with the updates of the Q-table. 


\subsection{Whittle Indices}



The Whittle Index policy assigns to each client a state-dependent scalar \(W_j(s_{j,r})\), which  captures the marginal utility of selecting client $j$ when it is in state $s_{j,r}$.
This scalar balances short-term rewards with long-term outcomes and is derived from a discounted single-agent Markov decision process (MDP) corresponding to the client.  

To align with the global objective in \eqref{eq:problem}, we define the immediate reward function as 
\begin{align}
& R_j(s_{j,r}, a_{j,r}) \nonumber \\
&\mathllap{=} - \mathbb{E}\left[ t^t_{j,r} + t^c_{j,r} + \lambda \left( F(\mathbf{w}^r) - F(\mathbf{w}^*) \right) \mid s_{j,r}, a_{j,r} \right]
\label{eq:reward_function}
\end{align}
where $\alpha_{j,r} \in (0, 1)$ indicates whether client $j$ is selected in round $r$. \(\lambda > 0\) is a balancing factor that reflects the relative importance of latency versus model accuracy. The reward quantifies the negative impact of scheduling a given client, incorporating both its local delay and its contribution to global loss reduction.




To define the Whittle Index, we introduce a subsidy $m$ associated with taking the passive action  \(a_{j,r} = 0\) and find the level at which the clients are indifferent between being selected or not. 


The value function \(V_j(s_{j,r}; m)\) satisfies the Bellman equation:
\begin{equation}
\begin{split}
    V_j(s_{j,r}; m) = \max_{a_{j,r} \in \{ 0, 1 \}} \Big\{ & R_j(s_{j,r}, a_{j,r}) + \mathbb{I}\{a_{j,r} = 0\} m \\
    & + \beta \sum_{y \in S}P^c_{a_{j,r}}(s_{j,r}, y) \cdot V_j(y; m) \Big\},
\end{split}
\label{eq:bellman_equation}
\end{equation}
where \(\beta \in (0, 1)\) is a discount factor, and \(\mathbb{I}\{\cdot\}\) denotes the indicator function. The transition probability \(P^c_{a_{j,r}}(s_{j,r}, y)\) denotes the state transition probability for class $c$ under action \(a_{j,r}\); specifically, $P^c_s(s_{j,r}, y)$ if active and $P^c_n(s_{j,r}, y)$ if passive.


The Whittle Index \(W_j(s_{j,r})\) is the smallest subsidy \(m\) for which the client is indifferent between being selected and not selected:
\begin{equation*}
\label{eq:whittle_index}
\begin{aligned}
    R_j(s_{j,r}, 0) + m &+ \beta \sum_{y \in S} P^c_n(s_{j,r}, y) \cdot V_j(y; m) \\
    &= R_j(s_{j,r}, 1) + \beta \sum_{y \in S}  P^c_s(s_{j,r}, y) \cdot V_j(y; m).
\end{aligned}
\end{equation*}

Under mild indexability assumptions, the set of states favoring the passive action expands monotonically with $m$, which implies that in each round, selecting the $|\mathcal{A}_r|$ clients with the highest $W_j(s_{j,r})$ yields a well-defined priority ordering. This balances the exploitation of clients in favorable states (e.g., normal) with the exploration of clients likely to become available.


If the transition probabilities \(P^c_{s}(x,y)\) and \(P^c_{n}(x,y)\) were known and client states $s_{j,r}$ observable, all Whittle indices \(W_j(s_{j,r})\) could, in principle, be precomputed. However, in federated settings, the server has no direct access to client states and can only infer them through observable quantities such as latency and selection feedback. Consequently, a static index policy is inapplicable, motivating the need for  a dynamic estimation strategy to approximate $W_j(s_{j,r})$ based on empirical rewards and inferred state transitions.

\subsection{Q-Learning}



For each client \(j \in \mathcal{C}_r\) at round \(r\), we define the Q-function \(Q_j(s_{j,r}, a_{j,r}; m)\) to encapsulate the expected long-term reward under a subsidy \(m\), aligning with the structure in \eqref{eq:bellman_equation}. Specifically, for state \(s_{j,r} \in S\) and action \(a_{j,r} \in \{0, 1\}\), the Q-function is:
\begin{equation}
\begin{split}
    Q_j(s_{j,r}, a_{j,r}; m) &= R_j(s_{j,r}, a_{j,r}) + \mathbb{I}\{a_{j,r} = 0\} m \\
&\quad + \beta \sum_{y \in S}P^c_{a_{j,r}}(s_{j,r}, y) \cdot V_j(y; m),
\end{split}
\end{equation}
with \(V_j(s_{j,r}; m) = \max_{a_{j,r} \in \{0, 1\}} Q_j(s_{j,r}, a_{j,r}; m)\). The subsidy \(m\) adjusts the trade-off between active and passive actions, which is critical for approximating \(W_j(s_{j,r})\). Since \(P^c_{a_{j,r}}(s_{j,r}, y)\) is unavailable to the server, Q-learning is used to iteratively update an estimated Q-function, \(\hat{Q}_{j,r}(s_{j,r}, a_{j,r}; m)\), based on observed outcomes rather than explicit transition probabilities.

The Q-learning recursion initializes \(\hat{Q}_{j,0}(s_{j,r}, a_{j,r}; m)\) arbitrarily 
for all \(s_{j,r} \in S\), \(a_{j,r} \in \{0, 1\}\), and \(m\) over a discrete candidate set of approximated Whittle indices \(\Lambda\). At each round \(r\), the server selects action \(a_{j,r} = \arg \max_{a \in \{0, 1\}} \hat{Q}_{j,r}(s_{j,r}, a; m)\) for a hypothesized state \(s_{j,r}\), observes the resulting latency \(t^t_{j,r} + t^c_{j,r}\) (capped at \(t_{\text{max}}\)), and infers the next state \(s_{j,r+1}\) indirectly via latency patterns or historical data. The update rule is:
\begin{equation*}
\begin{split}
\hat{Q}_{j,r+1}(s_{j,r}, a_{j,r}; m) &= (1 - \eta_r) \hat{Q}_{j,r}(s_{j,r}, a_{j,r}; m) +\\
 \eta_r \Big[ R_j(s_{j,r}, a_{j,r}) + \mathbb{I}\{a_{j,r} &= 0\} m  + \beta \max_{a \in \{0, 1\}} \hat{Q}_{j,r}(s_{j,r+1}, a; m) \Big],
\end{split}
\end{equation*}
where \(\eta_r \in [0, 1]\) is the learning rate, typically decreasing with \(r\) (e.g., \(\eta_r = r^{-1/2}\)), and \(\beta\) ensures convergence by discounting future rewards. This update refines \(\hat{Q}_{j,r}\) using empirical rewards, bypassing the need for  \(P^c_{s}(x,y)\) or  \(P^c_{n}(x,y)\), and applies only to the state-action pair \((s_{j,r}, a_{j,r})\) experienced at round \(r\); for other pairs, \(\hat{Q}_{j,r+1} = \hat{Q}_{j,r}\).

Convergence to the true \(Q_j\) is guaranteed under standard conditions: \(\sum_{r=0}^\infty \eta_r = \infty\) and \(\sum_{r=0}^\infty \eta_r^2 < \infty\)~\cite{floudas_neuro-dynamic_2008}, assuming sufficient state-action exploration. In practice, the server aggregates updates across clients within each class \(c \in C\), leveraging their shared Markovian dynamics to enhance learning efficiency.

\subsection{Federated Learning with WILF-Q}

We now introduce the Whittle Index Learning for Federated Q-learning (WILF-Q) algorithm, which is designed to dynamically approximate \(W_j(s_{j,r})\) online via Q-learning. Specifically, WILF-Q exploits the insight from \eqref{eq:whittle_index} that \(W_j(s_{j,r})\) is the \(m\) equating active and passive Q-values. This motivates a tailored Q-learning strategy, which approximates these indices online, enabling scalable client selection to minimize \(\mathcal{T}^\phi\) while meeting the accuracy constraint in \eqref{eq:problem}. 


We now briefly describe the key steps in WILF-Q. Recall from \eqref{eq:whittle_index} that \(W_j(s_{j,r})\) is the subsidy \(m\) at which the active and passive actions yield equivalent value, i.e.,
\begin{equation}
    Q_j(s_{j,r}, 1; m) - Q_j(s_{j,r}, 0; m) = 0.
\end{equation}

Given that \(P^c_{s}(x,y)\) and  \(P^c_{n}(x,y)\) cannot be directly observed by the server, we adopt Q-learning to estimate \(\hat{Q}_{j,r}(s_{j,r}, a_{j,r}; m)\) and derive \(\hat{W}_j(s_{j,r})\) as the value of \(m\) that minimizes the difference between respective Q-values. For each client \(j \in \mathcal{C}_r\) and state \(s_{j,r} \in S\), the estimated Whittle Index is
\begin{equation}
\label{Whittle Index}
\hat{W}_j(s_{j,r}) = \arg \min_{m \in \Lambda} |\hat{Q}_{j,r}(s_{j,r}, 1; m) - \hat{Q}_{j,r}(s_{j,r}, 0; m)|.
\end{equation}
This approximation is based on on iterative refinements of \(\hat{Q}_{j,r}\), as outlined in the previous subsection, using observed latencies \(t^t_{j,r} + t^c_{j,r}\) as empirical feedback.

The WILF-Q policy then operates by selecting the \(|\mathcal{A}_r|\) clients with the highest \(\hat{W}_j(s_{j,r})\) at each round \(r\). To ensure sufficient exploration across the subsidy space, we introduce an exploration probability \(\gamma_r \in [0, 1]\), which occasionally perturbs the selection process. With probability \(\gamma_r\), the action vector \(\mathbf{a}_r = (a_{j,r} : j \in \mathcal{C}_r)\) is randomly permuted, and \(\hat{W}_j(s_{j,r})\) is reassigned a uniform random value from \(\Lambda\). This mechanism prevents premature convergence to suboptimal indices, particularly for clients in less frequently observed states (e.g., busy).

The WILF-Q algorithm simultaneously updates \(\hat{Q}_{j,r}\) and \(\hat{W}_j(s_{j,r})\). For a client \(j\) at round \(r\), the Q-value update follows the recursion:
\begin{equation}
\begin{aligned}
&\hat{Q}_{j,r+1}(s_{j,r}, a_{j,r}; \hat{W}_j(s_{j,r})) \\
&= (1 - \eta_r) \hat{Q}_{j,r}(s_{j,r}, a_{j,r}; \hat{W}_j(s_{j,r})) \\
&\quad + \eta_r \left[ R_j(s_{j,r}, a_{j,r}) + \mathbb{I}\{a_{j,r} = 0\} \hat{W}_j(s_{j,r}) \right] \\
&\quad + \beta \max_{a \in \{0, 1\}} \hat{Q}_{j,r}(s_{j,r+1}, a; \hat{W}_j(s_{j,r})),
\end{aligned}
\label{Update Q-Function}
\end{equation}
where \(R_j(s_{j,r}, a_{j,r}) \) reflects both of the expected latency and a penalty on model inaccuracy, and \(\hat{W}_j(s_{j,r})\) substitutes \(m\) as the current estimation of the Whittle index. Note that updates occur only for the observed state-action pair \((s_{j,r}, a_{j,r})\); otherwise, \(\hat{Q}_{j,r+1} = \hat{Q}_{j,r}\).

This dual learning process enhances efficiency by focusing computational effort on index estimation rather than exhaustive Q-table maintenance. In addition, the server consolidates updates within each class, reducing redundancy and accelerating convergence. The detailed procedures of WILF-Q are presented in Algorithm~\ref{alg:WILF-Q}.

\begin{algorithm}[t]
    \KwRequire{Initial model $\mathbf{w}^0$, approximated subsidy set $\Lambda$, client set $\mathcal{C}_r$, learning rate $\{\eta_r\}$, discount factor $\beta$, accuracy weight $\lambda$, convergence threshold $\alpha$, exploration probability $\{\gamma_r\}$}
    \DontPrintSemicolon
    \SetAlgoLined
    \hrule
    \tcp{Initialization}
    \For{each client $j \in \mathcal{C}_r$}{
    Initialize $\hat{Q}_j(s,a;m)$ for all $s \in S$, $a \in \{0,1\}$, $m \in \Lambda$\;
    Randomly initialize $\hat{W}_j(s)$ for all $s \in S$
}
    $r \leftarrow 1$, $\text{converged} \leftarrow \text{False}$\;
    \While{\text{not converged}}{
        \tcp{Update estimated states and indices}
        \For{each client $j \in \mathcal{C}_r$}
        {
        Estimate current state $s_{j,r}$\;
        Update Whittle index using~\eqref{Whittle Index}
        }
        
        \tcp{Client selection}
        Select $|\mathcal{A}_r|$ clients with largest  $\hat{W}_j(s_{j,r})$ to form $\mathcal{A}_r$\;
        
        \If{with probability $\gamma_r$}{
            \tcp{Exploration: randomize estimated indices}
            $\hat{W}_j(s_{j,r}) \sim \mathcal{U}(\Lambda)\,\ \forall j \in \mathcal{C}_r$\;
        }
        \tcp{Client local training and update of Q-table}
        \For{each $j \in \mathcal{A}_r$}{
        Local update: compute $\mathbf{w}_j^{r}$ from $\mathbf{w}^{r-1}$\;
        Compute reward $R_j(s_{j,r}, a_{j,r})$ using~\eqref{eq:reward_function}\;
        Estimated next state $s_{j,r+1} \sim P(\cdot \mid s_{j,r}, a_{j,r})$\;

            \tcp{Update Q-table}
            Update approximated Q-table using~\eqref{Update Q-Function}\
        }
        
        \tcp{Model aggregation by FedAvg}
        $\mathbf{w}^{r} \leftarrow \sum_{j \in \mathcal{A}_r} k_j \mathbf{w}_j^{r}$
        
        \If{$[F(\mathbf{w}^{r}) - F(\mathbf{w}^*)] \leq \alpha$}{
            $\text{converged} \leftarrow \text{True}$, $R^\phi \leftarrow r$\;
        }
        $r \leftarrow r+1$\;
    }
    
    \KwRet{Final model $\mathbf{w}^{R^\phi}$}
    \caption{Whittle Index Learning in Federated Q-learning (WILF-Q)}
    \label{alg:WILF-Q}
\end{algorithm}

By approximating \(W_j(s_{j,r})\) in real-time, WILF-Q enables the server to adaptively prioritize clients that can improve accuracy within shorter periods of time. While the joint state space of the RMABP is exponential, the overall complexity of WILF-Q is approximately $O(R^{\phi}Ed)$, with the updates of $\hat{Q}$ and $\hat{W}$ adding only $O(|\mathcal{C}_r|)$ per round. The polynomial complexity guarantees the scalability of WILF-Q in large-scale scenarios.

\subsection{Convergence Analysis}

We now analyze the convergence performance under WILF-Q. Consider the global loss function $F(\mathbf{w})$ in~\eqref{eq:global_loss}. Under standard assumptions, we assume $F$ is $ L $-smooth and satisfies the Polyak–Łojasiewicz (PL) condition:
$\tfrac{\xi}{2}\|\nabla F(\mathbf w)\|^2 \le F(\mathbf w)-F(\mathbf w^*)$, for some constant $\xi > 0 $. At each round $r$, the server selects a fixed fraction $\rho$ of clients, broadcasts the current global model $\mathbf w^{\,r-1}$, and aggregates their local updates. Assuming local gradients are unbiased with bounded variance, we have
$\mathbb E \|g_{j,r} - \nabla F(\mathbf w^{r-1})\|^2 \le \sigma^2$.

The Q-table used for index estimation evolves according to the recursion in~\eqref{Update Q-Function} with step size $ \eta_r = \eta_0 / (r+1)^\theta $, where $\tfrac{1}{2} < \theta \le 1$. Since the controlled Markov chains are finite, irreducible, and aperiodic, it is ensured that:
\begin{equation}
    \| \mathbb E[\hat Q_{j,r} - Q_j^*] \|_\infty = \mathcal O(r^{-1/2}),
\quad
\hat Q_{j,r} \xrightarrow{\text{a.s.}} Q_j^*.
\end{equation}

Applying the Lipschitz property of the mapping in~\eqref{Whittle Index}, the estimated Whittle index satisfies
\begin{equation}
    \mathbb E\big[|\hat W_{j,r}(s) - W_j(s)|\big] = \mathcal O(r^{-1/2}),
\end{equation}
and the corresponding mis-ordering probability decays as $p_r = \mathcal O(r^{-1/2})$ by Hoeffding’s inequality. Let $\gamma_{\max} \triangleq \max_{j,s} \left| R_j(s,1) - R_j(s,0) \right|$, which upper-bounds the worst additional instantaneous cost incurred by selecting a sub-optimal client. Incorporating the effect of index errors into the FedAvg descent, we obtain
\begin{align}
    \mathbb E\big[F(\mathbf w^{r}) - F(\mathbf w^*)\big]
    &\le \Big(1 - \frac{\xi \rho \eta_r}{2}\Big)
        \mathbb E\big[F(\mathbf w^{r-1}) - F(\mathbf w^*)\big] \nonumber \\
    &\quad + L \sigma^2 \eta_r^2 + \eta_r \gamma_{\max} p_r.
\label{eq:convergence_analysis}
\end{align}

Substituting $\eta_r = \eta_0 / (r+1)$ and $p_r = \mathcal O(r^{-1/2})$, the residual terms scale as $\mathcal O(r^{-2})$ and $\mathcal O(r^{-3/2})$, both are summable over $r$. Telescoping ~\eqref{eq:convergence_analysis} from $r=1$ to $R^\phi$ yields the convergence result:
\begin{equation}
    \mathbb E\big[F(\mathbf w^{R^\phi}) - F(\mathbf w^*)\big] = \mathcal O(1/R^\phi).
\end{equation}
Here, all system-dependent parameters including local update frequency, model dimension and client heterogeneity, are absorbed into constant factors, and do not affect the asymptotic convergence rate.

\section{Experiments}

\subsection{Experimental Setup}
We consider a wireless network with $|\mathcal{C}_r| = 100$ clients. The uplink transmit power is fixed at $23\,\mathrm{dBm}$ for all clients, and the noise power is set to $\sigma_\gamma^2=10^{-5}\,\mathrm{W}$, which models the thermal noise level at the receiver during wireless transmission. 
Clients are grouped into three different classes based on normalized compute and connectivity: 30 high-capacity clients ($c_j^1\!\sim\!\mathcal{U}(0.7,1.0)$, $B_j\!=\!100$\,Mbps), 40 medium-capacity clients ($c_j^2\!\sim\!\mathcal{U}(0.4,0.7)$, $B_j\!=\!50$\,Mbps), and 30 low-capacity clients ($c_j^3\!\sim\!\mathcal{U}(0.2,0.4)$, $B_j\!=\!20$\,Mbps). 
In addition, each class has its own transition matrix. In general, high-performance clients (class 1) tend to persist in or recover to the normal state when selected or not selected, whereas lower-performance clients (class 3) remain longer in limited or busy states. 

We use the MNIST dataset and a small CNN tailored to MNIST, inspired by the ResNet9. The model is trained using Adam with a learning rate of $10^{-3}$ and a batch size of $32$ for both training and testing. Data heterogeneity across clients is introduced via a Dirichlet distribution with concentration parameter $\tau$. Small $\tau$ values (e.g., 0.1) induce higher non-IID levels or more skewed data distributions, while larger $\tau$ values  (e.g., 10) yield more balanced data. We use the subsidy set $\Lambda = \{0.1, 0.2, 0.3, 0.4, 0.5\}$, the discount factor $\beta = 0.9$, the exploration probability $\gamma_r =1/r$, and $\lambda$ chosen to balance delay and accuracy. A larger $\Lambda$ improves index resolution but slows learning; a decaying $\gamma_r$ ensures early exploration and late exploitation. We also set $\eta_r = r^{-1/2}$ and $\alpha = 0.15$.


\subsection{Baselines}
 We consider the following baseline policies:
 \begin{itemize}
 \item Random selection (RAN): $|\mathcal{A}_r|$ clients are selected uniformly at random each round.
 \item Efficiency-First (EF): $|\mathcal{A}_r|$ clients with the lowest expected latency are chosen for each round.
  \item Classical Q-Learning (CQL): Selects \(|\mathcal{A}_r|\) clients via standard Q-learning, with  \(\hat{Q}_{j,r}(s_{j,r}, a_{j,r})\) updated (without Whittle Indices) to minimize the latency.
  \item Upper Confidence Bound (UCB): $|\mathcal{A}_r|$ clients are selected via the UCB algorithm as proposed in~\cite{Chen2022heterogeneous}. It estimates each client's latency by combining its empirical mean and a confidence term that decays with selection frequency.
 \item Full-Information (FI):
$|\mathcal{A}_r|$ clients are selected based on perfect information of the Whittle Indices. Note that this only serves as a theoretical upper bound of performance, as the transition matrices are not known and thus the Whittle Indices could not possibly be computed in practice. 
\end{itemize}

\subsection{Numerical Results}



\begin{figure}[!t]
    \centering
    \includegraphics[width=0.95\columnwidth]{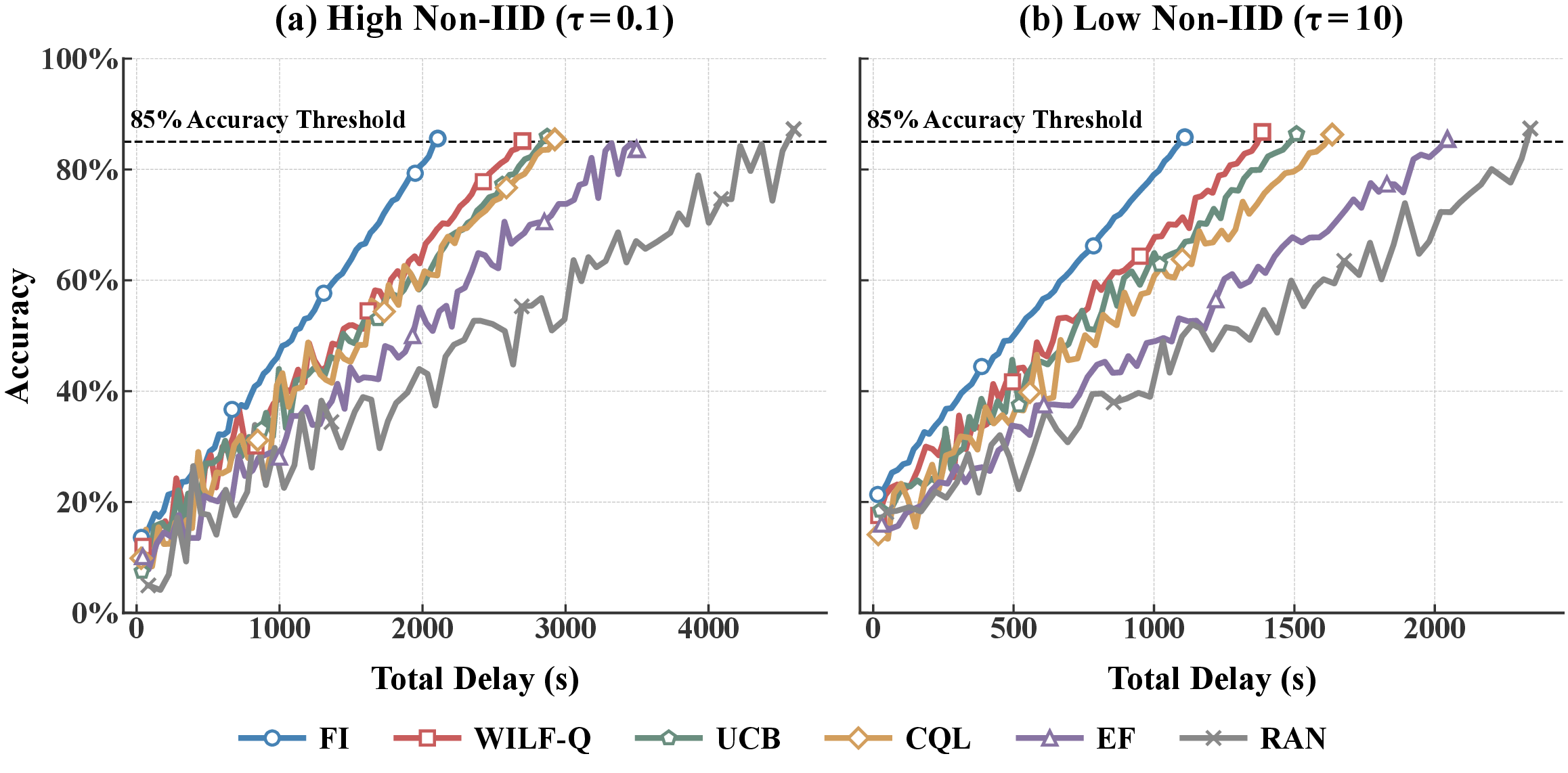}
    \caption{Comparison of accuracy vs. total delay for different client selection policies under two Non-IID settings. (a) High Non-IID case ($\tau=0.1$). (b) Low Non-IID case ($\tau=10$).}
    \label{fig:accuracy_comparison}
\end{figure}

\begin{figure}[!t]
\centering
\includegraphics[width=\columnwidth]{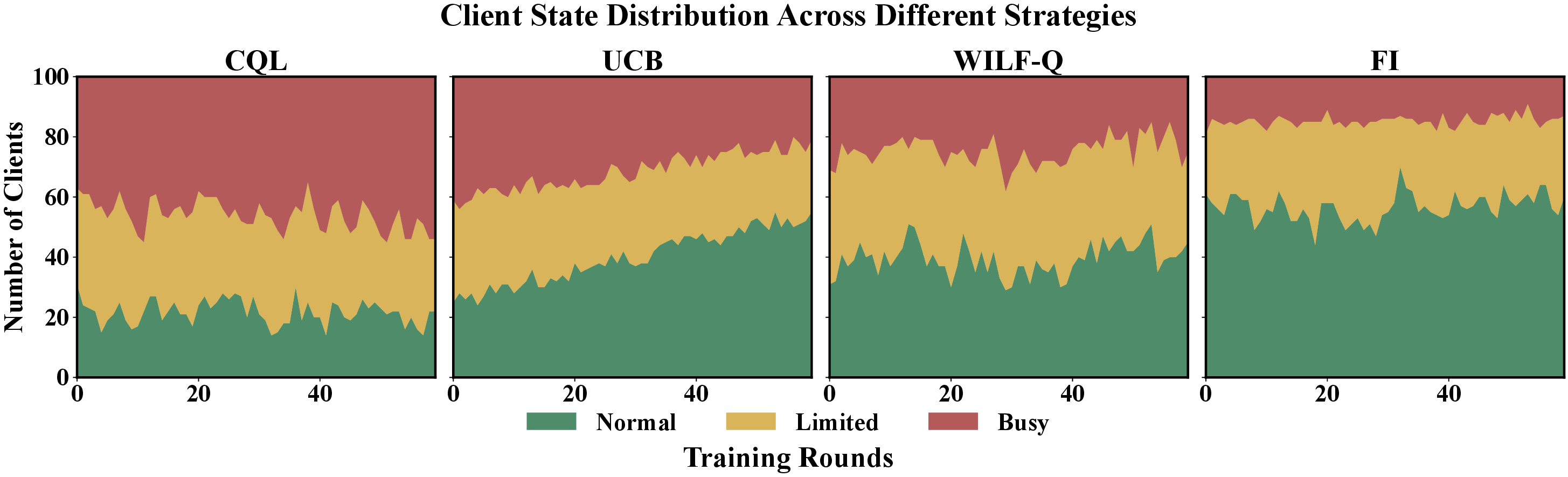}
\caption{Client state distribution over training rounds: CQL vs UCB vs WILF-Q vs FI ($\tau = 0.1$).}
\label{fig:client_state_distribution}
\end{figure}

All simulation results presented in this section have a $95\%$ confidence interval within $3\%$ of the observed mean.

Fig.~\ref{fig:accuracy_comparison} presents the relationship between total delay and testing accuracy under high ($\tau = 0.1$) and low ($\tau = 10$) non-IID settings, respectively. The results show that, WILF-Q achieves the closest performance in terms of learning efficiency to the theoretically optimal FI and outperforms all other benchmarks. Also, while both WILF-Q and CQL show significant fluctuations in learning accuracy in the early exploration stages of Q-learning, WILF-Q stabilizes more rapidly and achieves higher accuracy in fewer rounds compared to CQL, effectively demonstrating the advantage of selecting clients via the Whittle-index. Specifically, under $\tau = 0.1$, WILF-Q reduces the training time by 41\%, 21\%, 10\% and 7\% compared to RAN, EF, CQL and UCB, respectively. When $\tau = 10$, WILF-Q outperforms RAN, EF, and CQL by 45\%, 29\%, 9\% and 3\%, respectively.
Another observation is that, in the high non-IID case, all strategies experience slower convergence and higher variance in both accuracy and delay. 



Fig.~\ref{fig:client_state_distribution} shows the client state distributions under CQL, UCB, WILF-Q and FI across training rounds with $\tau = 0.1$. 
The state distributions under WILF-Q are very similar to those under FI, reflecting the effectiveness of WILF-Q in accurately learning the optimal client-selection policy. UCB lies between WILF-Q and CQL, by optimistic exploration, it steadily increases the normal state portion. Moreover, compared with CQL, WILF-Q achieves a significantly more balanced distribution of client states, with fewer clients remaining in the busy state. This indicates that the Whittle index-based approach effectively manages dynamic client availability throughout the FL process. 


\section{Conclusion}
We proposed WILF-Q, a Whittle index-based Q-learning approach that effectively addresses client selection challenges in wireless FL systems under dynamic and non-IID conditions. We showed that WILF-Q can adapt to evolving client states, balancing accuracy and delay without requiring explicit knowledge that may affect the learning efficiency, including state transitions and client data distributions. By unifying state-aware learning from historical rewards, our proposed approach has shown superior performance in terms of convergence efficiency compared to benchmark policies through numerical experiments. 

\section*{Acknowledgment}
This work is partly supported by the Guangdong Provincial/Zhuhai Key Laboratory of 
Interdisciplinary Research and Application for Data Science, Project 2022B1212010006 and in part by Guangdong Higher Education Upgrading
Plan (2021-2025) UIC [R0400001-22] and [R0400024-22].


%





\ifCLASSOPTIONcaptionsoff
  \newpage
\fi

\bibliographystyle{IEEEtran}
\bibliography{IEEEabrv,Bibliography}

\begin{thebibliography}{10}
\providecommand{\url}[1]{#1}
\csname url@rmstyle\endcsname
\providecommand{\newblock}{\relax}
\providecommand{\bibinfo}[2]{#2}
\providecommand\BIBentrySTDinterwordspacing{\spaceskip=0pt\relax}
\providecommand\BIBentryALTinterwordstretchfactor{4}
\providecommand\BIBentryALTinterwordspacing{\spaceskip=\fontdimen2\font plus
\BIBentryALTinterwordstretchfactor\fontdimen3\font minus
  \fontdimen4\font\relax}
\providecommand\BIBforeignlanguage[2]{{%
\expandafter\ifx\csname l@#1\endcsname\relax
\typeout{** WARNING: IEEEtran.bst: No hyphenation pattern has been}%
\typeout{** loaded for the language `#1'. Using the pattern for}%
\typeout{** the default language instead.}%
\else
\language=\csname l@#1\endcsname
\fi
#2}}
\renewcommand\BIBentryALTinterwordstretchfactor{4}

\bibitem{yang_federated_2019}
Q.~Yang, Y.~Liu, T.~Chen, and Y.~Tong, ``Federated {Machine} {Learning}:
  {Concept} and {Applications},'' \emph{ACM Trans. Intell. Syst. Technol.},
  vol.~10, no.~2, pp. 12:1--12:19, Jan. 2019.

\bibitem{qi2025energy}
H.~Qi, J.~Luo, Z.~Xu, Q.~Li, J.~Yin, and J.~Wu, ``Energy efficient power
  control for over-the-air federated learning in satellite communications,''
  \emph{IEEE Commun. Lett.}, vol.~29, no.~7, pp. 1530--1534, 2025.

\bibitem{fu_client_2023}
L.~Fu, H.~Zhang, G.~Gao, M.~Zhang, and X.~Liu, ``Client {Selection} in
  {Federated} {Learning}: {Principles}, {Challenges}, and {Opportunities},''
  \emph{IEEE Internet Things J.}, July 2023.

\bibitem{nishio_client_2019}
T.~Nishio and R.~Yonetani, ``Client {Selection} for {Federated} {Learning} with
  {Heterogeneous} {Resources} in {Mobile} {Edge},'' in \emph{{ICC} 2019 - 2019
  {IEEE} {International} {Conference} on {Communications} ({ICC})}, May 2019,
  pp. 1--7.

\bibitem{abdulrahman_fedmccs_2021}
S.~Abdulrahman, H.~Tout, A.~Mourad, and C.~Talhi, ``{FedMCCS}: {Multicriteria}
  {Client} {Selection} {Model} for {Optimal} {IoT} {Federated} {Learning},''
  \emph{IEEE Internet Things J.}, vol.~8, no.~6, pp. 4723--4735, Mar. 2021.

\bibitem{Arouj2022}
A.~Arouj and A.~M. Abdelmoniem, ``Towards energy-aware federated learning on
  battery-powered clients,'' in \emph{Proc. of FedEdge}, 2022, pp. 7--12.

\bibitem{Whittle1988}
P.~Whittle, ``Restless bandits: activity allocation in a changing world,''
  \emph{J. Appl. Probab. Stat.}, vol.~25, no.~A, p. 287–298, 1988.

\bibitem{albelaihi_green_2022}
R.~Albelaihi, L.~Yu, W.~D. Craft, X.~Sun, C.~Wang, and R.~Gazda, ``Green
  {Federated} {Learning} via {Energy}-{Aware} {Client} {Selection},'' in
  \emph{2022 {IEEE} {Global} {Communications} {Conference}}, Dec. 2022, pp.
  13--18.

\bibitem{wu_towards_2023}
C.~Wu, M.~Xiao, J.~Wu, Y.~Xu, J.~Zhou, and H.~Sun, ``Towards {Federated}
  {Learning} on {Fresh} {Datasets},'' in \emph{2023 {IEEE} 20th {International}
  {Conference} on {Mobile} {Ad} {Hoc} and {Smart} {Systems} ({MASS})}, Sept.
  2023, pp. 320--328.

\bibitem{Chen2022heterogeneous}
S.~Chen, X.~Wang, P.~Zhou, W.~Wu, W.~Lin, and Z.~Wang, ``Heterogeneous
  semi-asynchronous federated learning in {Internet of Things}: A multi-armed
  bandit approach,'' \emph{IEEE Trans. Emerg. Top. Comput.}, vol.~6, no.~5, pp.
  1113--1124, 2022.

\bibitem{maciel_federated_2024}
F.~Maciel, A.~M. de~Souza, L.~F. Bittencourt, L.~A. Villas, and T.~Braun,
  ``Federated learning energy saving through client selection,''
  \emph{Pervasive Mob. Comput.}, vol. 103, p. 101948, Oct. 2024.

\bibitem{Lee2018Speeding}
K.~Lee, M.~Lam, R.~Pedarsani, D.~Papailiopoulos, and K.~Ramchandran, ``Speeding
  up distributed machine learning using codes,'' \emph{IEEE Trans. Inf.
  Theory}, vol.~64, no.~3, pp. 1514--1529, 2018.

\bibitem{Zhao2022System}
Z.~Zhao, J.~Xia, L.~Fan, X.~Lei, G.~K. Karagiannidis, and A.~Nallanathan,
  ``System optimization of federated learning networks with a constrained
  latency,'' \emph{IEEE Trans. Veh. Technol.}, vol.~71, no.~1, pp. 1095--1100,
  2022.

\bibitem{Papadimitriou1994complexity}
C.~H. Papadimitriou and J.~N. Tsitsiklis, ``The complexity of optimal queueing
  network control,'' in \emph{Proc. of IEEE 9th Annual Conference on Structure
  in Complexity Theory}, 1994, pp. 318--322.

\bibitem{floudas_neuro-dynamic_2008}
D.~P. Bertsekas, ``Neuro-{Dynamic} {Programming}: {NDP},'' in
  \emph{Encyclopedia of {Optimization}}, C.~A. Floudas and P.~M. Pardalos,
  Eds.\hskip 1em plus 0.5em minus 0.4em\relax Boston, MA: Springer US, 2008,
  pp. 2555--2560.

\end{thebibliography}

\vfill


\end{document}